\documentclass[conference,a4paper]{IEEEtran}
\IEEEoverridecommandlockouts
\usepackage{cite}
\usepackage{amsmath,amssymb,amsfonts}
\usepackage{algorithmic}
\usepackage{graphicx}
\usepackage{textcomp}
\usepackage{xcolor}
\usepackage{multicol}
\usepackage{siunitx, makecell}
\usepackage{mathtools}
\usepackage{mathptmx} 
\usepackage{subfigure}
\usepackage{amsbsy}
\usepackage{bm}
\usepackage{url}
\usepackage{accents}

\makeatletter
\let\old@ps@headings\ps@headings
\let\old@ps@IEEEtitlepagestyle\ps@IEEEtitlepagestyle
\def\confheader#1{%
	\def\ps@IEEEtitlepagestyle{
		\old@ps@IEEEtitlepagestyle
		\def\@oddhead{\strut\hfill#1\hfill\strut}
		\def\@evenhead{\strut\hfill#1\hfill\strut}
	}
	\ps@headings
}
\makeatother
\confheader{
	\small{Proceedings of the 11th RSI International Conference on Robotics and Mechatronics (ICRoM 2023), Dec. 19-21, 2023, Tehran, Iran} 
}
\usepackage{eso-pic}
\newcommand{\placetextbox}[3]{
	\setbox0=\hbox{#3}
	\AddToShipoutPictureFG*{ \put(\LenToUnit{#1\paperwidth},\LenToUnit{#2\paperheight}){\vtop{{\null}\makebox[0pt][c]{#3}}}
	}
}
\placetextbox{.23}{0.055}{\textbf{\small{979-8-3503-0810-5/23/\$31.00~\copyright 2023 IEEE}}}

\usepackage{hyperref}
\hypersetup{
	colorlinks=true,
	linkcolor=black,
	citecolor=black,
	filecolor=black,
	urlcolor=black,
}
\def\BibTeX{{\rm B\kern-.05em{\sc i\kern-.025em b}\kern-.08em
		T\kern-.1667em\lower.7ex\hbox{E}\kern-.125emX}}

\makeatletter 
\newcommand{\linebreakand}{%
\end{@IEEEauthorhalign}
\hfill\mbox{}\par
\mbox{}\hfill\begin{@IEEEauthorhalign}
}
\makeatother 

\begin{document}
	
	\title{
		Design and Fabrication of a Conceptual Model for a Portable Ring Shaped Medical Imaging Apparatus
	}
	
	\author{\IEEEauthorblockN{Mahdi Nozari}
		\IEEEauthorblockA{\textit{Biomechatronics and Cognitive Engineering Research Lab} \\
			\textit{School of Mechanical Engineering} \\
			\textit{Iran University of Science and Technology}\\
			nozarim@ut.ac.ir}
		\and
		\IEEEauthorblockN{MohammadSajad Sokot}
		\IEEEauthorblockA{\textit{Biomechatronics and Cognitive Engineering Research Lab} \\
			\textit{School of Mechanical Engineering} \\
			\textit{Iran University of Science and Technology}\\
			s-sokot@mecheng.iust.ac.ir}
		\and
				\linebreakand 
		\IEEEauthorblockN{Arshia Sheikh Hasani}
		\IEEEauthorblockA{\textit{Biomechatronics and Cognitive Engineering Research Lab} \\
			\textit{School of Mechanical Engineering} \\
			\textit{Iran University of Science and Technology}\\
			arshia.sheikhhassani@student.hs-anhalt.de}
		\and

		\IEEEauthorblockN{Borhan Beigzadeh}
		\IEEEauthorblockA{\textit{Biomechatronics and Cognitive Engineering Research Lab} \\
			\textit{School of Mechanical Engineering} \\
			\textit{Iran University of Science and Technology}\\
			b\_beigzadeh@iust.ac.ir}
	}


\maketitle

\begin{abstract}
Medical imaging plays a crucial role in various healthcare procedures, with the demand for efficient and precise imaging methods steadily increasing. One such critical application is spinal surgery, where the sensitivity of the operation necessitates imaging solutions that minimize patient movement while ensuring accuracy and convenience. Due to this, portable imaging devices are getting more and more relevant. Although several portable imaging devices exist for capturing images from different angles, a majority of them still rely on manual operation and lack the capability to rotate a full 360 degrees. In this research, a conceptual model of an apparatus was designed and fabricated that can easily take images from all 360 degrees with its motorized axes and can be controlled remotely, So it will alleviate the harmful effects of X-rays for the surgeon and the medical staff .The acquired 360-degree images are subsequently transformed into 3D models using advanced 3D image reconstruction techniques. 72 images were taken with the apparatus and reconstructed into a 3D model using the SURF method and with an error of around 7$\%$. This platform can be used in further research such as online surgery verification, robotic surgery integration and also motion artifact compensation using its motorized axes.
\end{abstract}


\section{Introduction}
Neurosurgery is one of the most common and delicate operations worldwide. An estimate of 13.5 million operations are needed each year that consist of different types, such as craniotomy and spinal cord deformity correction, and only half of them are performed \cite{b1}. Given the delicate nature of surgeries involving the brain and spinal cord, the presence of devices that enhance precision and efficacy in these procedures becomes crucial. The typical procedure for these operations requires medical images before and after the operation, respectively, to be used for planning and validation. By using a portable imaging apparatus, the need to move the patient to the imaging room will be eliminated, which will lead to an increase in efficiency and precision. To help with this, another method that was developed to help solve the issues is using a 2D fluoroscopy device known as the C-arm \cite{b2}. C-arm revolutionized medical imaging technology, and as a result, operations were done with much more efficiency. Nowadays, motorized and more advanced C-arm systems help surgeons take 3D images and can be controlled remotely.\\
Surgical robots have emerged as a significant topic of research. Particularly, telesurgery has caught many researchers’ attentions in recent years. These systems need to be properly controlled, both from position coordination \cite{Yazdankhoo2022_1} and safe interaction \cite{Yazdankhoo2022_2} perspectives. The pivotal criteria in controller design should encompass stability \cite{Yazdankhoo2022_3} and transparency \cite{Nikpour2021,Yazdankhoo2022_4}, which will result in the system operating properly.\\
As robotic surgery is getting more relevant, it is starting to get used in operation to lower the cost and increase accuracy \cite{b3,b4}. An appropriate imaging device will decrease the risk of surgery furthermore, if it can provide the robot with 3D models, specially in operations with a high risk of error\cite{b5,b6}. However, in less developed countries, the conventional method is used as there are an insufficient number of portable devices available and even fewer motorized ones. This issue and the lack of capability to rotate 360 degrees have led to the development of more advanced devices, such as an O-Arm. The new devices will be able to rotate completely, making them more versatile, and have more degrees of freedom to aid with the imaging procedure\cite{b7}. The risk of exposure of the medical staff to x-ray is decreased in overall as well\cite{b8,b9}.\\
Another important application of a motorized imaging device is the capability of compensation for body movements such as breathing, which causes artifacts in the final images\cite{b10}. To reduce the effect of breathing on the images, a breathing model can be developed\cite{b11} and then be given to a predictive control scheme, so the axes of the device can move and eliminate the movements as much as possible.
In this research, a device is proposed to take images automatically as a proof of concept for the mechanical mechanisms and also the 3D reconstruction algorithm. The device is not needed to be full-scale, as it is not yet going to be used with an X-ray tube. This setup must have the following modules:
\begin{itemize}
	\item \textbf{Motorized axes: }\\
	An automatic imaging device needs to be motorized so it can accept movement commands. Not only do the motorized axes give the operator precise control over the positioning of the device, but they also add the ability to be remotely controlled, resulting in less harmful X-ray exposure for the doctors and medical staff who are in the most danger. The assessed x-ray exposure of the O-arm device has shown that this control over the movement of the axes allows further advanced algorithms to be tested and developed.
	\item \textbf{Imaging Module: }\\
	In a fluoroscopy imaging device, a tube will generate x-rays, which a detector on the other side of the object receives. It is important to keep in mind that working with x-rays requires extra care and expertise. Given this fact, it is easier and more convenient to use a normal RGB camera to acquire the 360-degree images to develop the algorithm and modify it with certain methodology\cite{b12} later so it can also work with other cameras or even fluoroscopy images.
	\item \textbf{Incomplete gantry ring: }\\
	As the device is aimed at being a portable apparatus to be used inside the operation room, there must exist a mechanism that can be used to open up the ring so the ring can be placed around the ring.
	\item \textbf{Online 3D Reconstruction: }\\
	As mentioned earlier, the areas to which a portable imaging device can be applied require a real-time 3D model of the object. For this purpose, the 3D reconstruction algorithm must be run as the images are taken to aid the operation or be given as references to a robotic surgery platform. In order for this to happen, a connection from the microcontroller of the device to a computer must be made to receive the commands.
\end{itemize}
Surface from Motion (SfM) is a computer vision technique used to reconstruct the three-dimensional structure of objects or scenes from a sequence of two-dimensional images or video frames. Several key methods have been developed to facilitate the process of SfM. Three notable ones are Scale-Invariant Feature Transform (SIFT), Harris Corner Detection, and Speeded-Up Robust Features (SURF). We will introduce these methods briefly.
\begin{itemize}
	\item SIFT:\\
	SIFT is a widely used feature extraction and matching technique in SfM. It operates by identifying keypoints or interest points in images that are invariant to scale, rotation, and illumination changes. SIFT extracts distinctive feature descriptors at these keypoints, which can be matched across multiple images to establish correspondences. This enables the computation of camera poses and the reconstruction of the 3D scene.
	\item Harris Corner Detection:\\
	This method is a quick and basic method for finding points in images. It shows the areas in the image where there are large changes in intensity in all directions. These points are usually located at corners or connecting points in the image. In SfM, the Harris corners of multiple images are detected and aligned to determine the alignment required for 3D reconstruction.
	\item SURF:\\
	SURF is another feature detection description method used in SfM. It is designed to be computationally efficient and robust to various image transformations. SURF identifies keypoints and extracts feature descriptors that are particularly well-suited for real-time applications.
\end{itemize}
To choose an optimal reconstruction method, the desired application has to be taken into account. In an online imaging device, it is clear that to update the 3D model, a huge computational load is exerted on the processing unit. Unlike some other feature detection and description methods, SURF is designed to be computationally efficient, making it well-suited for real-time applications and large-scale image datasets\cite{b13}. Moreover, because of the robustness of the method to image altercations, images with less quality are also viable in this case \cite{b14}.
\section{Design}
The design is laid out in different phases based on the sections of the device. This sections are then designed with details, stress analysis is done to prevent any failure or large unwanted deflections and when all components are placed, the actuators are chosen so they can carry the load they have to deal with.  
\begin{figure}[htbp]
	\centering
	{\includegraphics[width=0.7\linewidth]{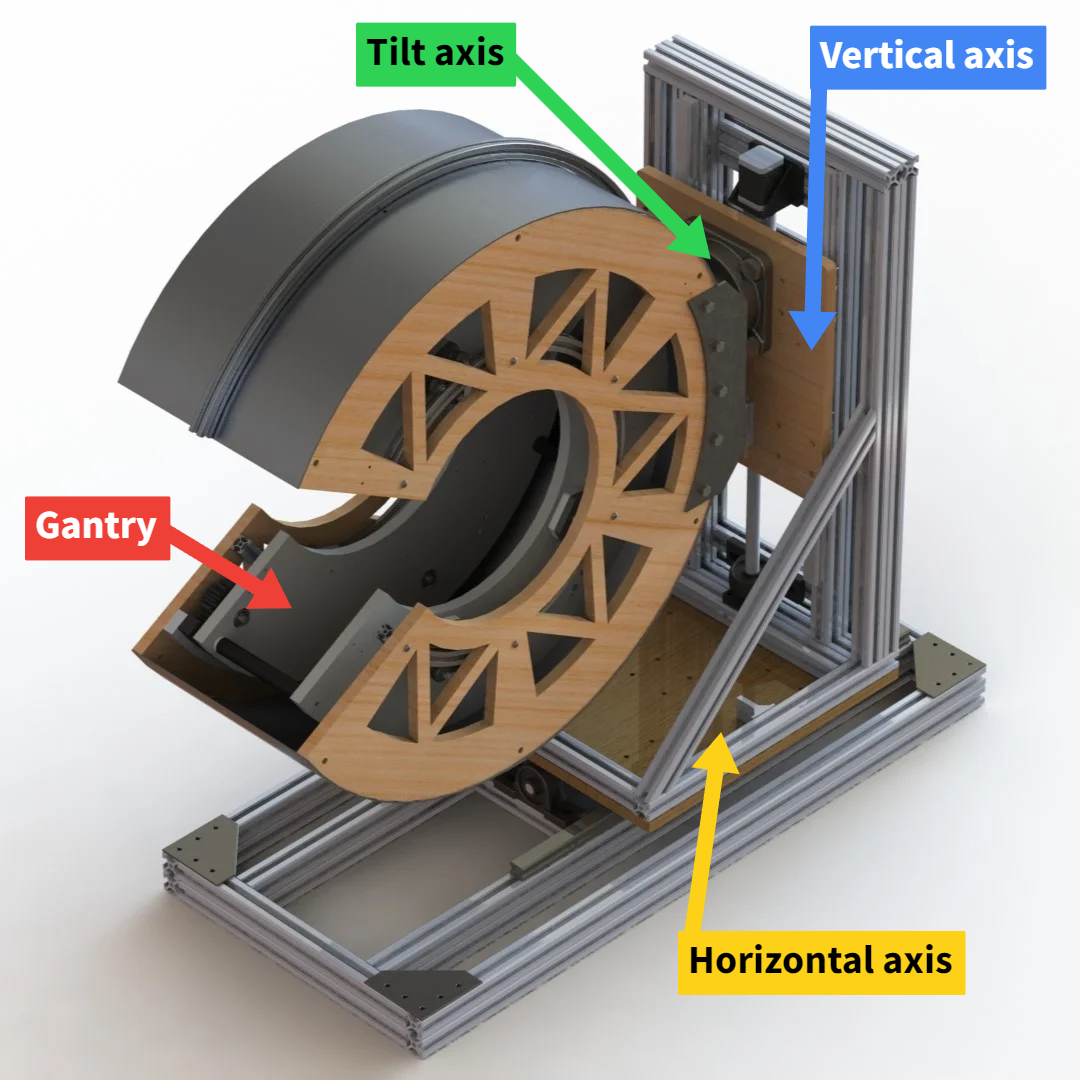}}
	\caption{Overall design of the ring shaped imaging device}
	\label{DesignOverall}
\end{figure}
\begin{figure*}[!t]
	\centering
	\subfigure[]{\includegraphics[height=0.21\linewidth]{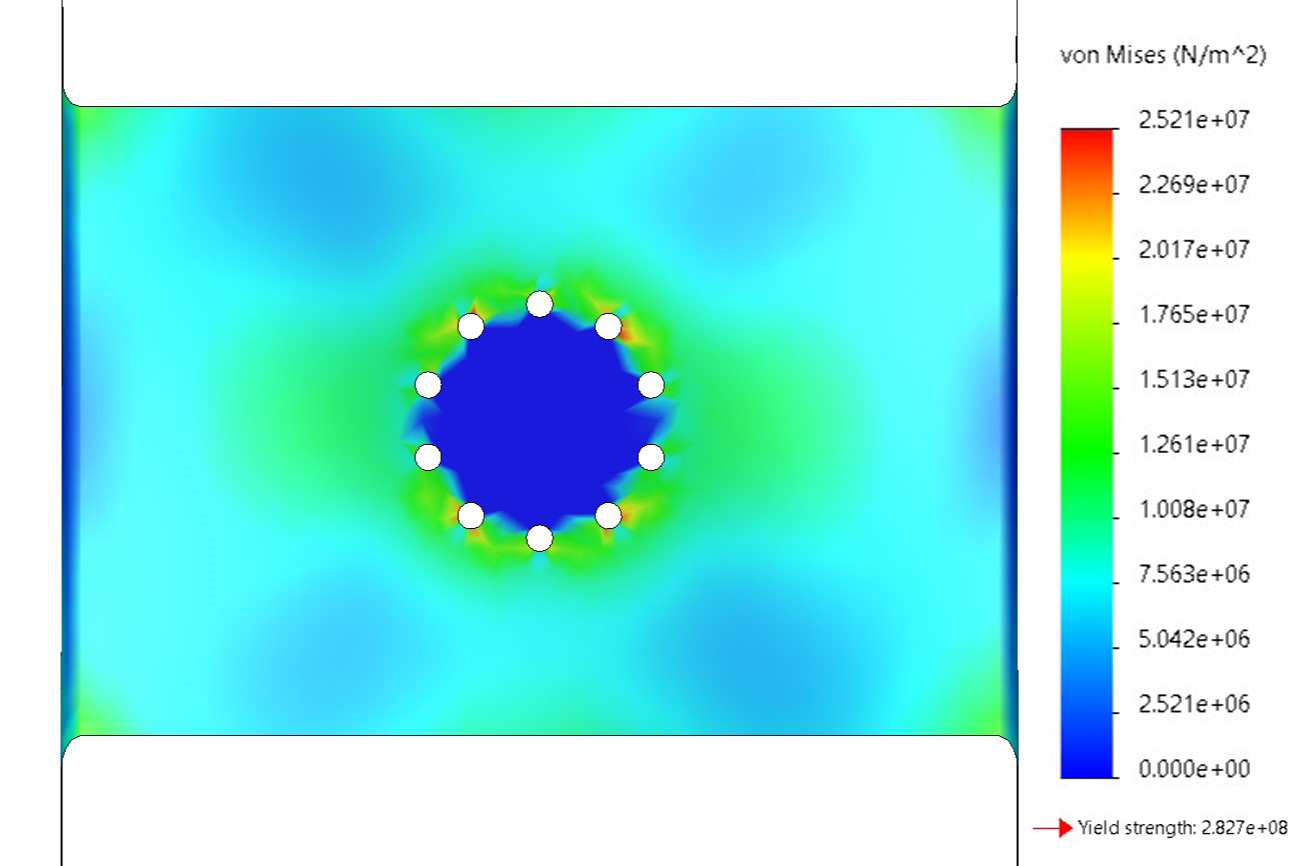}}
	\subfigure[]{\includegraphics[height=0.21\linewidth]{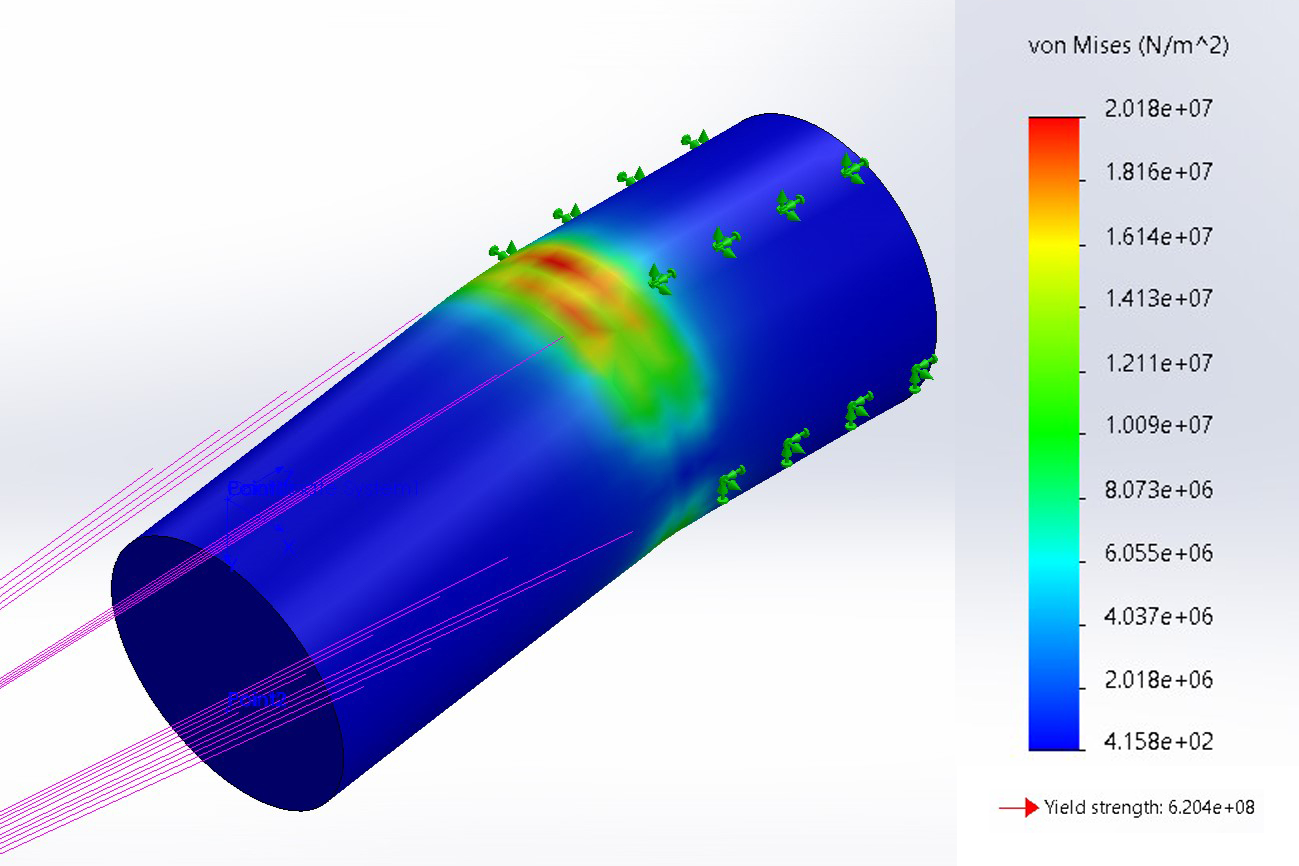}}
	\subfigure[]{\includegraphics[height=0.21\linewidth]{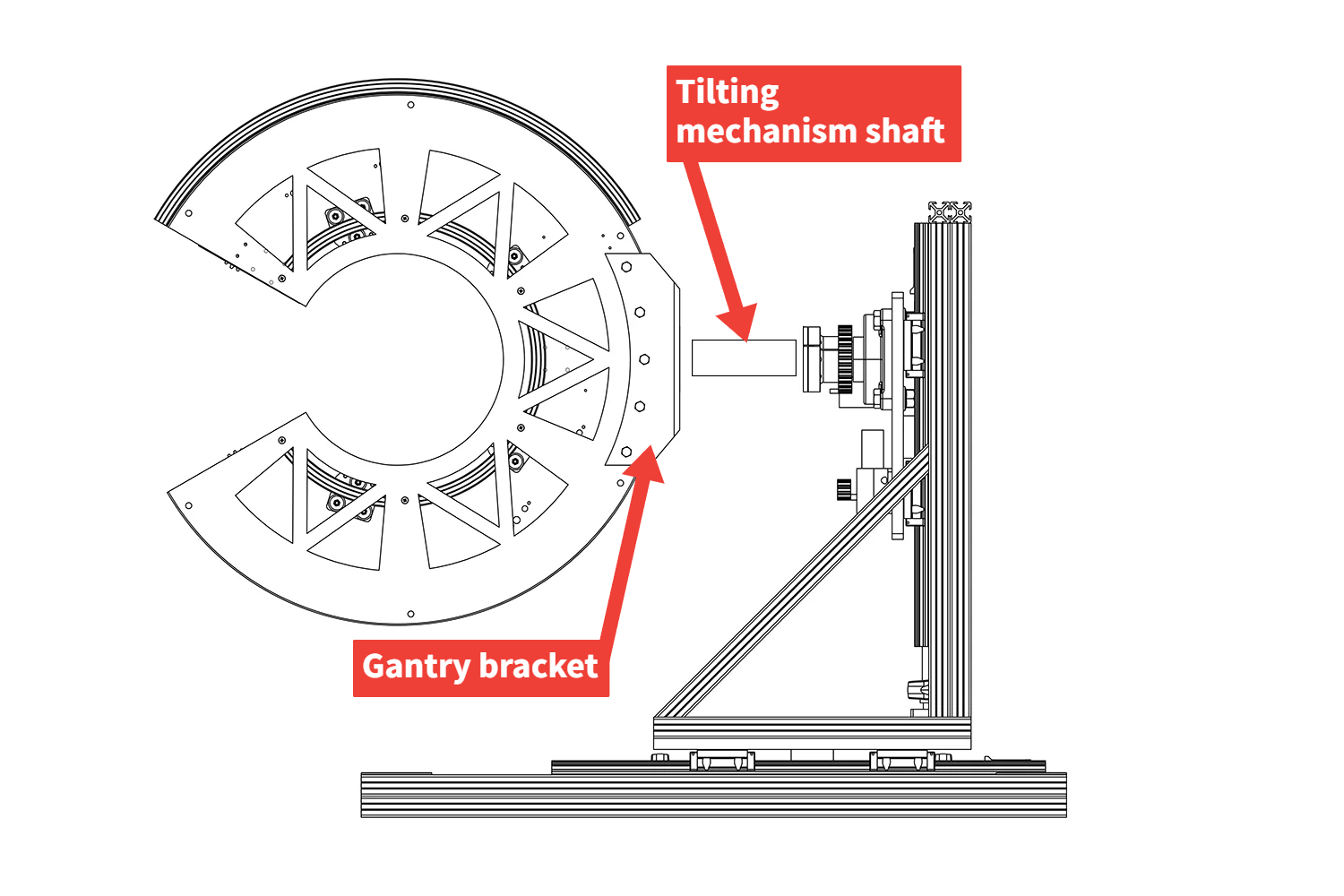}}
	\caption{FEM analysis (a)gantry bracket (b)tilting mechanism shaft (c)position of parts}
	\label{ShaftFEM}
\end{figure*}
\subsection{Overall Structure}
The device has two main sections, the cart and the gantry. The cart holds all the degrees of freedom that the device needs to adjust the gantry position. For this research 3 degrees are chosen, a horizontal axis, a vertical axis and the tilting axis which can be seen in Fig.~\ref{DesignOverall}. Because of the weight of the device and the nature of the loads which include moment, rail guides are the optimum choice for the linear axes. These axes are driven by leadscrews as they don't need speed and tinier motors can be used as the force is multiplied. For the tilting axis, a large bearing holds the gantry on only one side. This action is also motorized and can compensate for the angle of the operating table.
\subsection{Gantry Design}
The gantry of a medical imaging device houses all the imaging equipment and in some devices it can rotate around the object to take images from different angles. In this design the main challenge is to rotate the gantry while a section of it is missing. Because of this reason, a bearing cannot be used and curve guides are the solution. The drive system consists of an internal gear and pinions which are evenly distributed to further help the curve guides and keep the gantry in place. 
\begin{figure}[htbp]
	\centerline{\includegraphics[width=0.9\linewidth]{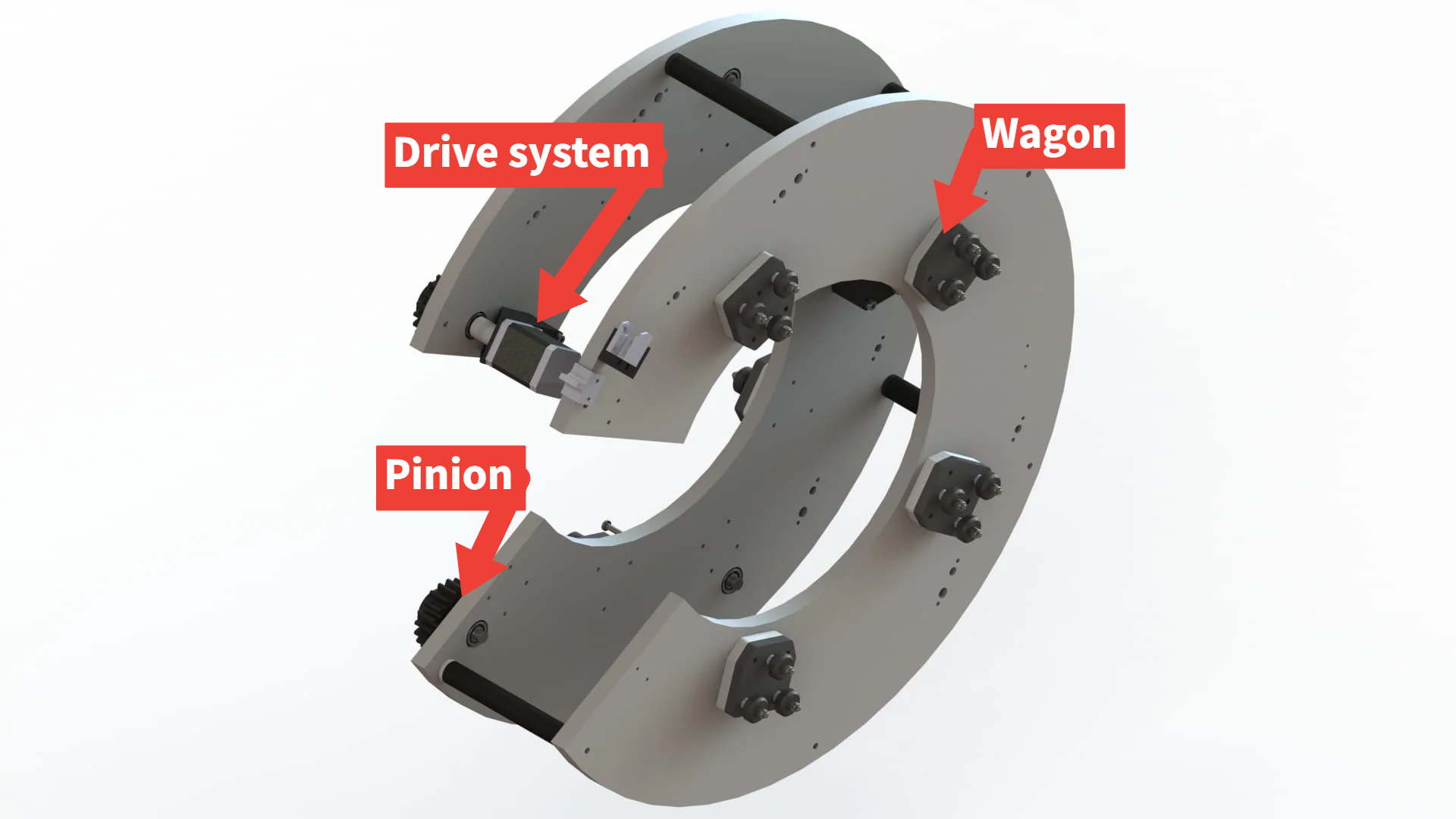}}
	\caption{Gantry of the device and its components}
	\label{Gantry}
\end{figure}
Given the fact that the ring misses a section, the rotor is unbalanced in nature. As the rotational velocity is in the range of $[0.05, 0.08] rad/s$, the force created by the rotor imbalance is of little concern. However, if the center of mass is not located in the center of the rotor, the actuators will have to be stronger to raise the CoM when it is lower than the center. By adding counterbalance weights, the actuators needed for the rotation will act better.\\
The designed gantry can be seen in Fig.~\ref{Gantry}.It is $200mm$ wide and has a diameter of $650mm$. Consisting of two planes connected with spacers which leaves the middle part empty for the imaging equipment to be installed, it allows objects up to the size of a sphere with a diameter of $300mm$. On both sides, wagons of the curved guide can be seen and only on one side, the drive system exists. Additional parts such as the cable management duct is also installed on the gantry.
\subsection{FEM Analysis}
Based on the overall design of the structure, the gantry of the device is somewhat like a cantilever beam, which is supported on only one side. based on the application of the device, adding another support on the other side of the gantry is not possible. This fact makes the supporting section of the device, the most critical part of it. To prevent this section from unwanted deflection of even failure, A static FEM analysis is carried out to determine the appropriate dimensions.\\
As the tilting mechanism exists in this section, a large shaft is inserted into a bearing, which allows rotation of the gantry. This shaft is then connected to a bracket by a powerlock, which keeps the shaft from rotating with respect to the gantry. The shaft and the bracket are most prone to failure while loaded. The analysis is presented in Fig.~\ref{ShaftFEM} with the properties explained in Table~\ref{Properties}. The exerted load on the parts in the analysis corresponds to the total gantry weight that these parts have to bear. The parts have been designed to have a reasonable amount of deflection and a high safety factor in case of failure. A trial and error approach was used to find a sufficient design. The parameters that were altered in the simulations were the diameter for the shaft, and the thickness for the bracket.
In the final design, the bracket and the shaft endure a maximum von-Mises stress of respectively $2.521e07 \space N/m^2$ and $2.018e07 \space N/m^2$ which results in safety factors 11.21 and 30.

\begin{table}[htbp]
	\caption{FEM analysis properties}
	\begin{center}
		\begin{tabular}{|>{\raggedright\arraybackslash}p{2cm}|>{\raggedright\arraybackslash}p{2cm}|>{\raggedright\arraybackslash}p{2cm}|}
			\hline
			\textbf{Property} & \textbf{Tilt Shaft} & \textbf{Gantry Bracket}\\
			\hline
			Type & Static & Static\\
			\hline
			Load Type & Remote load & Remote Load\\
			\hline
			Load & $400N$ & $400N$\\
			\hline
			Material & 1020 Steel & 1023 Carbon Steel Sheet\\
			\hline
			Young Modulus & $2.1e11 \space N/m^2$ & $2.1e11 \space N/m^2$\\
			\hline
			Poisson Ratio & $0.28$ & $0.29$\\
			\hline
			Yield Strength & $2.1e11 \space N/m^2$ & $2.827e11 \space N/m^2$\\
			\hline
			Maximum Stress (von-Mises) & $2.018e07 \space N/m^2$ & $2.521e07 \space N/m^2$\\
			\hline
			Maximum Displacement & $1.86e-6 \space m$ & $2.99e-6 \space m$\\
			\hline
		\end{tabular}
		\label{Properties}
	\end{center}
\end{table}
\begin{figure*}[htbp]
	\centering
	\subfigure[]{\includegraphics[height=0.16\textheight]{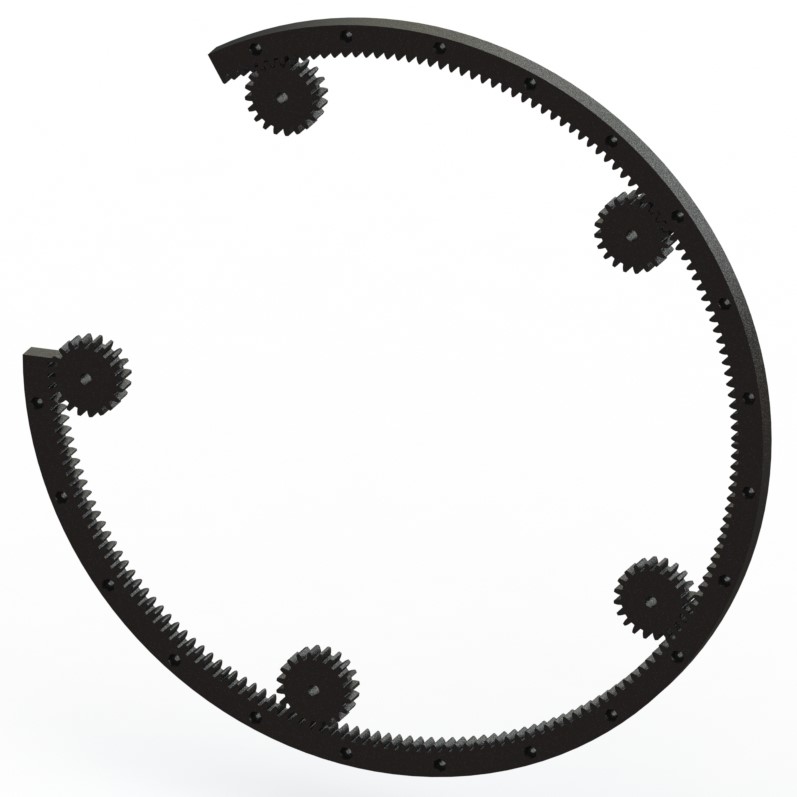}}
	\subfigure[]{\includegraphics[height=0.16\textheight]{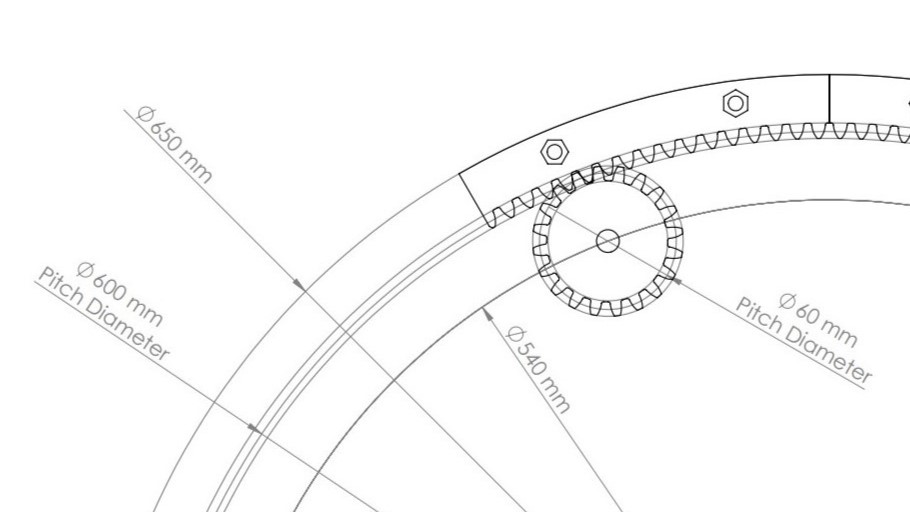}}
	\subfigure[]{\includegraphics[height=0.16\textheight]{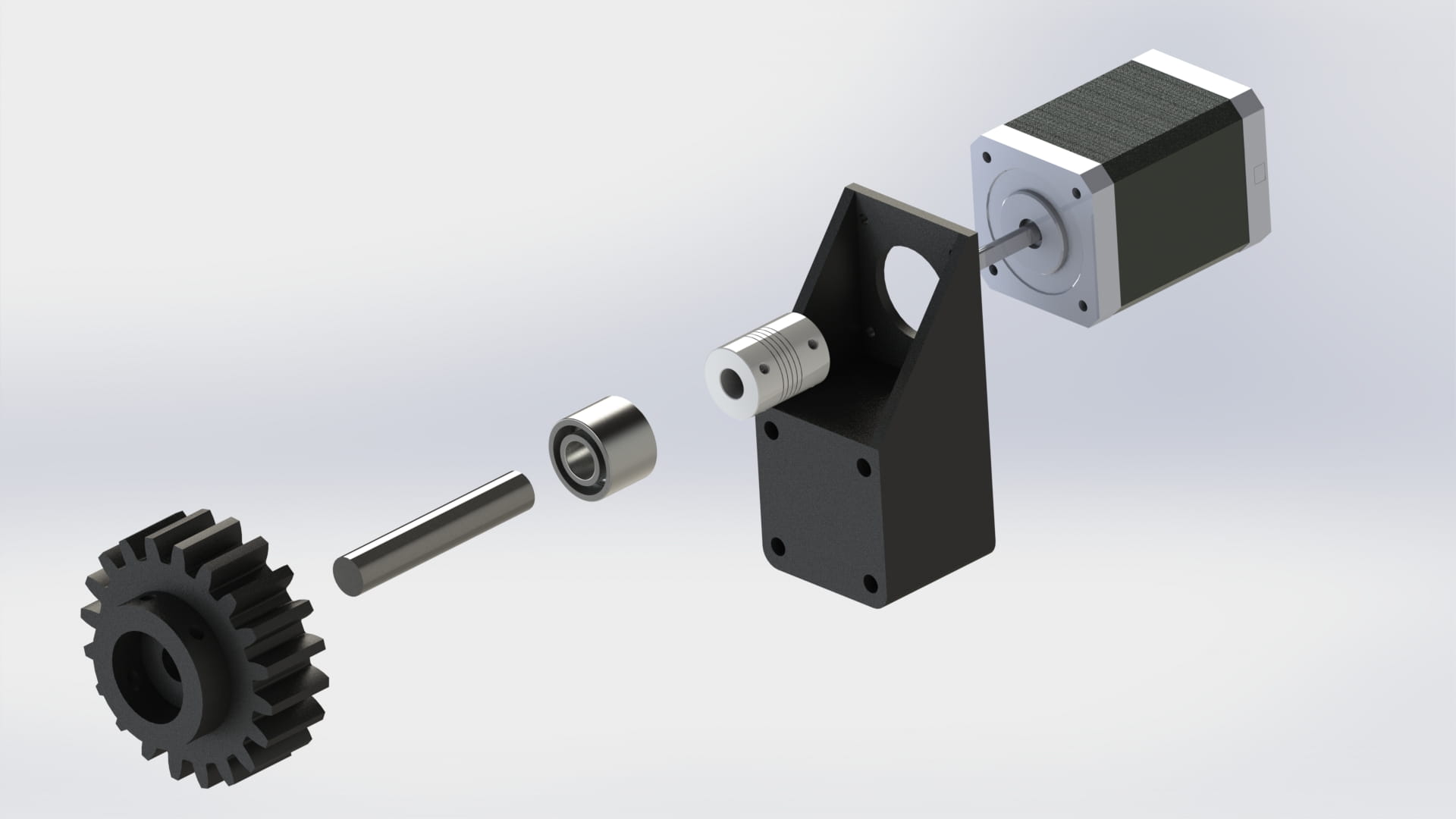}}
	\caption{Drive system (a) internal gear system (b) gear system dimensions (c) motor coupled with pinion}
	\label{Gears}
\end{figure*}
\subsection{Actuator Selection}
\begin{figure}[htbp]
	\centerline
	{\includegraphics[width=0.8\linewidth]{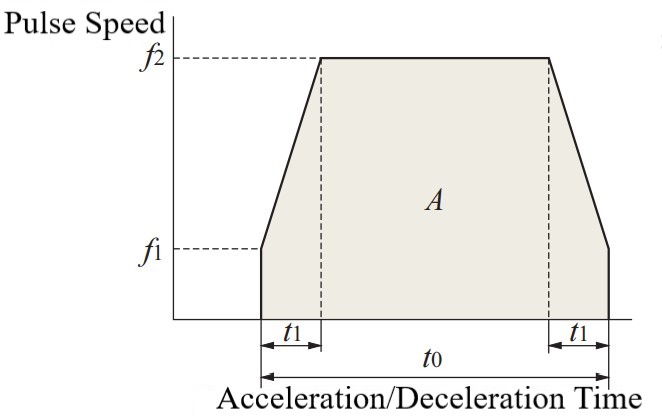}}
	\caption{Stepper motor acceleration/deceleration motion profile ($t_1$ is the acceleration time, $f_1$ is the starting pulse speed and $f_2$ is the operating pulse speed)}
	\label{AccDec}
\end{figure}

As mentioned earlier, five pinions are present in the gear system which can be seen in Fig.~\ref{Gears} . Two of these pinions are driven by motors. This will increase the amount of torque applied to the system and also keep the gantry in rotation even if one of the driven pinions exits the internal gear. The pinions are coupled directly to the motor using a flexible coupling. The gear ratio of the drive system is 10:1.\\
Stepper motors are chosen as the actuators of the horizontal and vertical axes, and also the gantry itself. To compute the needed torque for the required acceleration based on the load, the following equation is used.
\begin{equation}
	T_M = (T_L + T_a) \times S_f
\end{equation}
where $T_M$ is the required motor torque, $T_L$ is the load torque, $T_a$ is the acceleration torque and $S_f$ is the safety factor. The stepper motors operate with an acceleration/deceleration motion profile, which is depicted in Fig.~\ref{AccDec} \cite{b15}. $T_L$ is calculated with respect to the type of drive system, which for a leadscrew system is as below.
\begin{equation}
	T_L = (\frac{F P_B}{2 \pi \eta})
\end{equation}
where $\mu$ is the friction coefficient of the sliding surface, $\eta$ is the efficiency of the leadscrew, $P_B$ is the leadscrew lead and $F$ is the force in the moving direction which is calculated as below.
\begin{equation}
	F = F_A + mg(\sin(\theta) + \mu \cos(\theta))
\end{equation}
where $F_A$ is the external force on the carriage and $\theta$ is the angle of inclination. The explained torque is only enough to oppose the static load. So if friction is neglected (which isn't far from reality because of the rail guides), for a horizontal ($\theta=0 \deg$) carriage $F=0$ so $T_L=0$ and for a vertical ($\theta= 90 \deg$) carriage $F=mg$. To make the load move, The motor has to overcome the inertia of the load as well. The torque needed for this purpose is called the acceleration torque and as the name presents, it accelerates the carriage. It is calculated as below \cite{b4}.
\begin{equation}
	T_a = (j_0 \times i^2 + j_L) \times \frac{\pi \times \theta_s}{180} \times \frac{f_2 - f_1}{t_1}
\end{equation}
where $j_0$ is the rotor inertia, $j_L$ is the load inertia, $i$ is the gear ratio, $\theta_s$ is the step angle of the motor and $t_1$, $f_1$, $f_2$ are explained in Fig.~\ref{AccDec}.\\
After the mentioned calculations, a $5 kg.cm$ motor for the horizontal axis, and a $30 kg.cm$ motor (with gearbox) for the vertical axis have been chosen based on the parameters of the system.
\begin{figure*}[t]
	\centering
	{\includegraphics[width=0.8\linewidth]{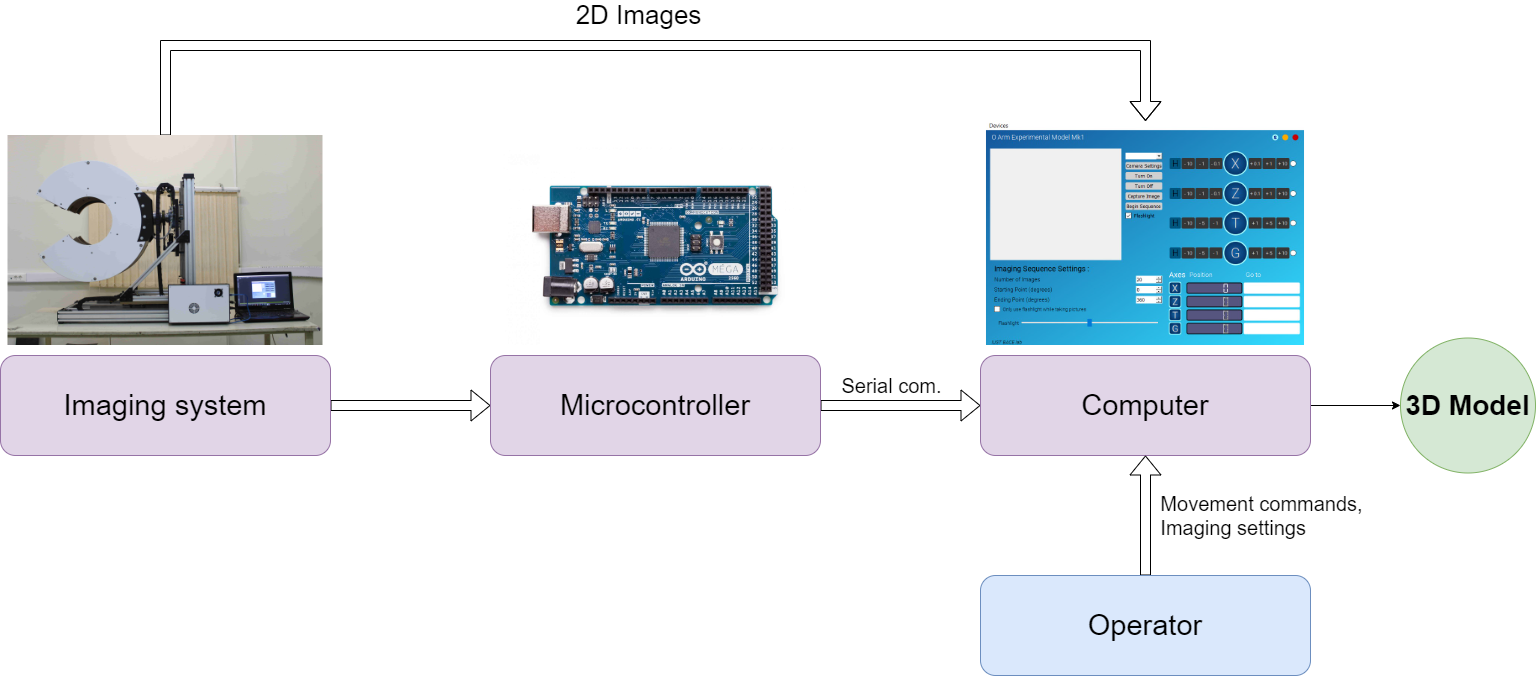}}
	\caption{3D model generation diagram}
	\label{Diagram}
\end{figure*}
\begin{figure*}[t]
	\centering
	\subfigure[]{\includegraphics[height=0.15\textheight]{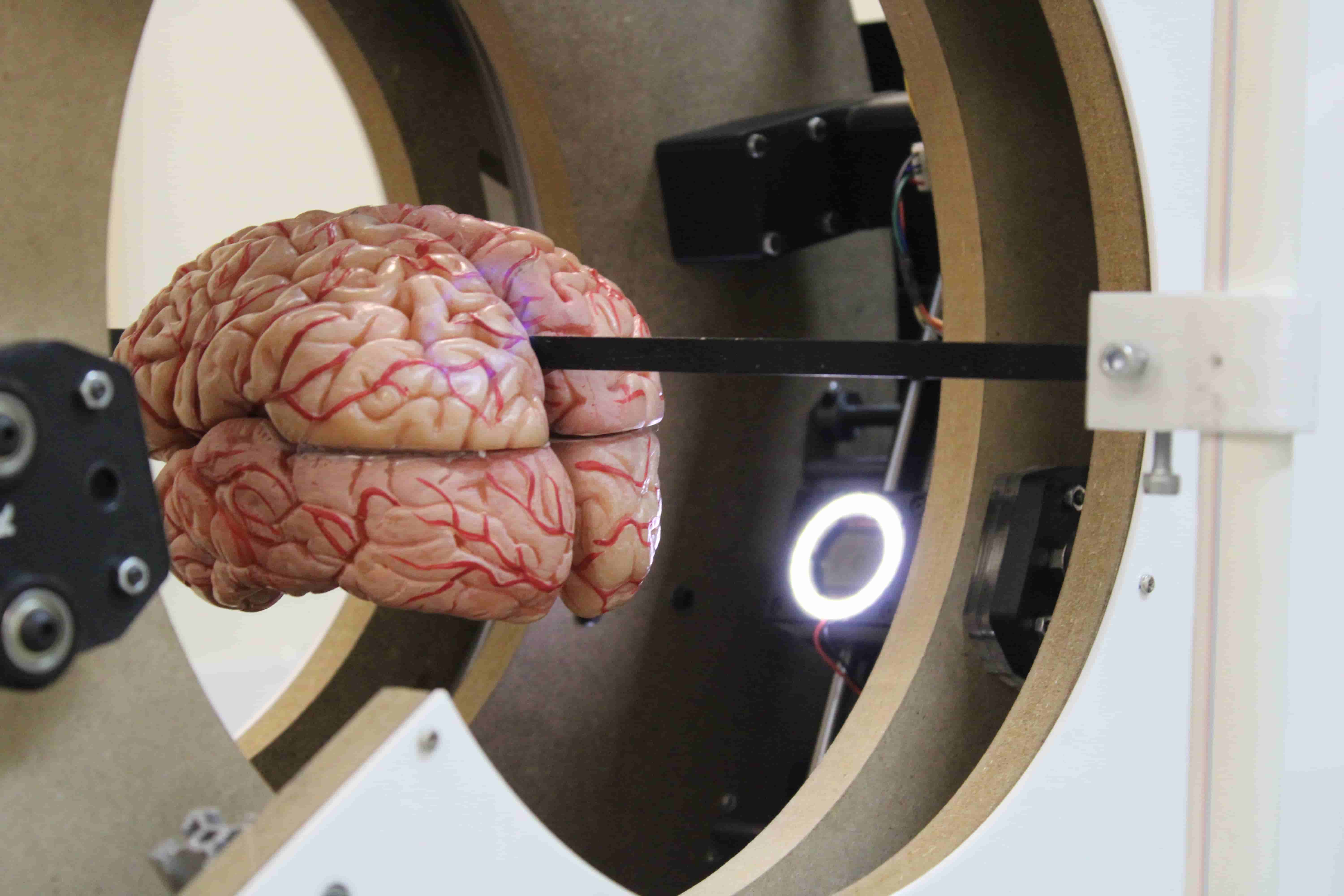}}
	\subfigure[]{\includegraphics[height=0.15\textheight]{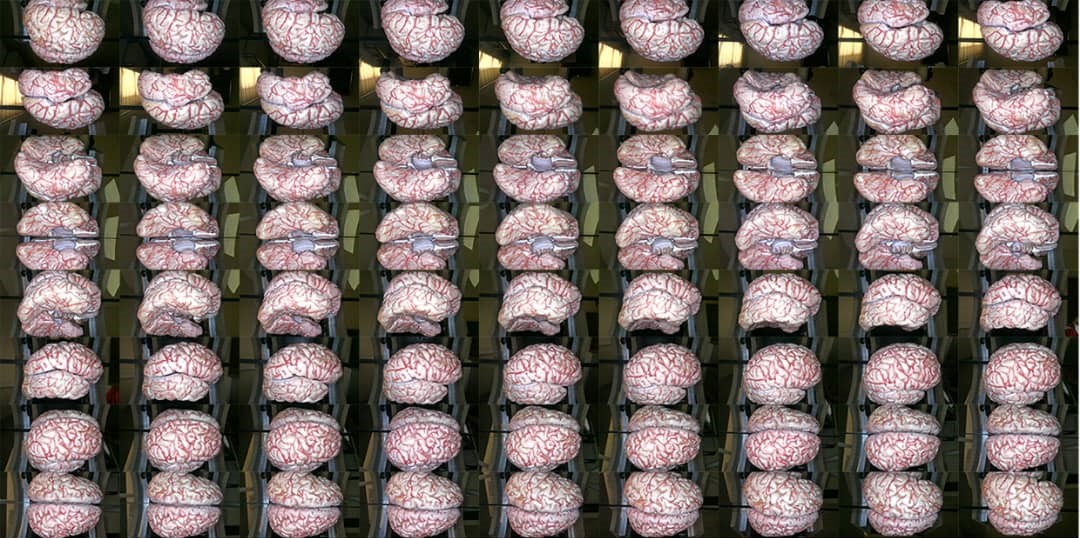}}
	\subfigure[]{\includegraphics[height=0.15\textheight]{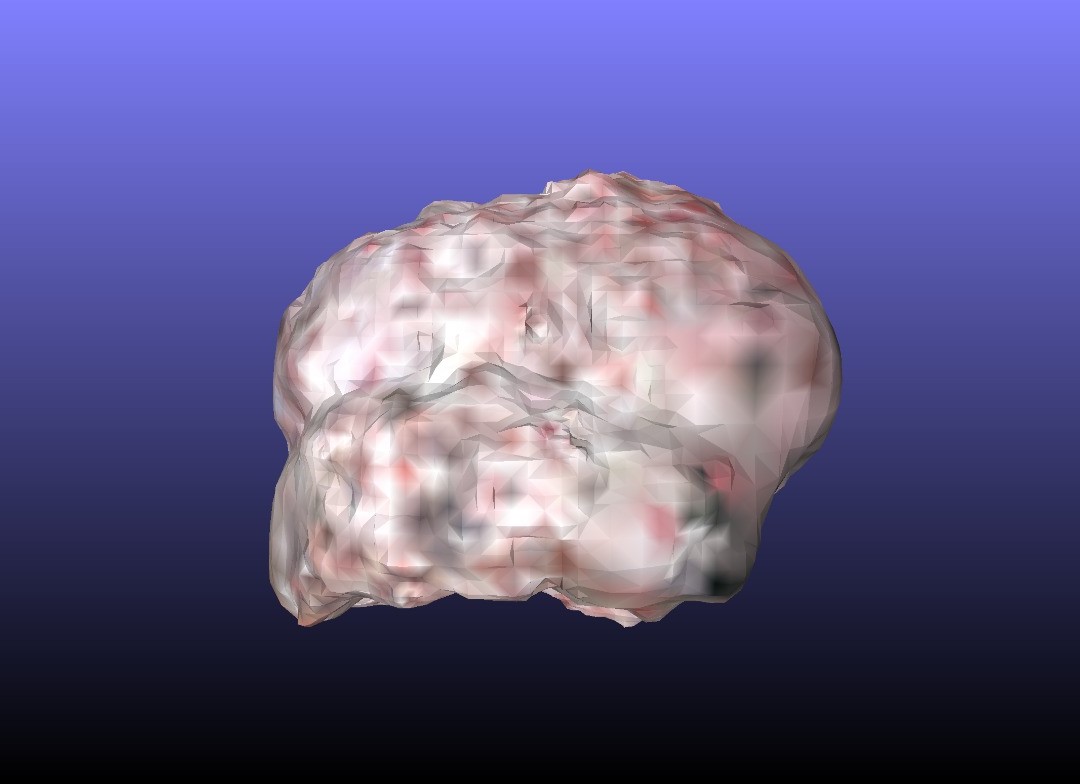}}
	\caption{Experimental test results (a) Object placement (b) 360-degree images - 72 images in total (c) Extracted 3D model}
	\label{Brain}
\end{figure*}
\section{3D reconstruction}
The SURF method uses a feature detector based on the Hessian matrix, which relies on integral images for computation time reduction, therefore called "Fast-Hessian" detector. The Integral The entry of an integral image at $(x, y)$ represents the sum of all pixels in the input image $I$ of a rectangular region formed by the point $(x, y)$ and the origin \cite{b3}, which is shown below.
\begin{equation}
	I(x, y)=\sum_{i=0}^x \sum_{j=0}^y I(i, j) \\
\end{equation}
For the point $\mathbf{x}=(x,y)$ inside an image, the Hessian matrix $\mathcal{H}$ at scale $\sigma$ is defined as follows
\begin{equation}
	\mathcal{H}(\mathbf{x}, \sigma)=\left[\begin{array}{ll}
		L_{x x}(\mathbf{x}, \sigma) & L_{x y}(\mathbf{x}, \sigma) \\
		L_{x y}(\mathbf{x}, \sigma) & L_{y y}(\mathbf{x}, \sigma)
	\end{array}\right]
\end{equation}
where $L_{xx}(\mathbf{x}, \sigma)$ is the convolution of the Gaussian 2nd order derivative $\frac{\partial^2}{\partial x^2} g(\sigma)$. By calculating the mentioned matrix, this method is able to detect the sudden changes around a certain point to find features like corners. The points which have high changing rates are considered as key points. These key points are then used to detect the features around them.
\section{Results}
The setup comprises of the mechanical system, the electronic system and microcontroller, a computer and the imaging camera which is mounted on the mechanical setup. The device operator can give movement commands or alter the imaging settings with a GUI on a computer. The computer will then send commands to the microcontroller through a serial connection and receive the images from a 2MP RGB camera. These images are inputted to the SURF algorithm and a point cloud file is extracted. A surface is then stretched on the point cloud and the 3D model will be ready. A summary of this procedure is shown in Fig.~\ref{Diagram}.

To test the mentioned procedure, a model of a human brain was placed in the gantry and a total of 72 images (each $5 \deg$) were taken. A view of The final 3D model and images are available in Fig.~\ref{Brain}. The physical model was $25.67cm$ wide where the extracted model had a width of $23.83cm$. Therefore an error of $1.84cm\approx7\%$ was present which due to the quality of the images and is reasonable.

\section{Conclusion}
In this research, by investigating different portable medical imaging devices, a conceptual model of an imaging device was developed that is capable of being used inside the operating room. Alongside the design, analysis, and fabrication of this device, a 3D reconstruction algorithm was coupled with the system to convert the 2D images into a 3D model. The system was able to successfully capture images from all sides autonomously and output a 3D model with reasonable error. However, due to the inaccessibility and danger of x-ray devices, an RGB camera was used. Nevertheless, the research proves the concept and can later be applied to a full-scale device as well, which is definitely the future direction of this project.

\IEEEtriggeratref{11}

\end{document}